\documentclass{article}
\usepackage[preprint]{neurips_2026} 

\usepackage[utf8]{inputenc}
\usepackage[T1]{fontenc}
\usepackage{hyperref}
\usepackage{url}
\usepackage{amsfonts}
\usepackage{nicefrac}
\usepackage{microtype}

\usepackage{todonotes}
\usepackage{makecell}
\usepackage{multirow}
\usepackage{booktabs}
\usepackage[table,xcdraw]{xcolor}
\usepackage{array}
\usepackage{caption} 

\newcolumntype{L}[1]{>{\raggedright\arraybackslash}p{#1}}

\begin{document}

\title{Emerging Flexible Designs for Geospatial Multimodal Foundation Models}

\author{%
  Philipe Dias \\
  Oak Ridge National Laboratory \\
  USA \\
  \And
  Waqwoya Abebe \\
  Oak Ridge National Laboratory \\
  USA \\
  \And
  Abhishek Potnis \\
  Oak Ridge National Laboratory \\
  USA \\
  \And
  Aristeidis Tsaris \\
  Oak Ridge National Laboratory \\
  USA \\
  \And
  Dan Lu \\
  Oak Ridge National Laboratory \\
  USA \\
  \And
  Xiao Wang \\
  Oak Ridge National Laboratory \\
  USA \\
  \And
  Dalton Lunga \\
  Oak Ridge National Laboratory \\
  USA \\
}

\maketitle

\begin{abstract}
\textbf{Abstract.} Foundation models are rapidly transforming Earth observation by enabling scalable pretraining across diverse unlabeled geospatial modalities. However, their architectural diversity—ranging from encoder-only to encoder-decoder and masked autoencoding paradigms—makes it challenging to assess performance trade-offs in a consistent manner. In this work, we present an apples-to-apples comparison of leading FM architectures designed for geospatial multimodal reasoning, with a particular focus on flexibility across varied spectral band configurations. We standardize pretraining using identical self-supervised learning objectives and training datasets, and evaluate all models under consistent parameterization on the GEOBench benchmark across classification and segmentation tasks. Our results offer new insights into the design trade-offs between model flexibility, modality alignment, and downstream task performance. By highlighting architectural strengths and limitations under controlled conditions, this study provides practical guidance for building next-generation geospatial foundation models capable of robust multimodal reasoning.

\end{abstract}

\section{Introduction}
As foundation models increase their dominance in interpreting Earth observation (EO) imagery so is the pursuit to design domain agnostic and transferrable architectures that trains with vast unlabeled data. Such architectures are preliminarily showing the potential to interpret vast amount of multifaceted EO data and promising improved human understanding of complex environmental phenomenon and impacts, including natural hazards, climate monitoring, and crop yield prediction. 

Powered by self-supervised learning methods the remote sensing applications are demonstrating the benefits of foundation models in parsing and exploitation of EO imagery. In contrast to their progress in single EO modality analysis, EO foundation models advancement to handle multisensor, multispectral and multiresolution EO data types is still lagging. In particular, enabling the following two desirable properties remains a big open challenge: i) ability to extract informative features from multiple modalities, instead of over-relying on predominant modalities (e.g., visible region of the spectrum); ii) enable usage in downstream scenarios where only a subset of pretraining modalities is present and when novel sensing modalities are introduced.

While the research community has recently introduced multiple new architectural schemes targeting these gaps, a systematic comparison between their design choices is still missing. Experimental protocols vary in terms of one or more variables: pretext task/objective, pretraining duration; finetuning strategy, downstream tasks, datasets, and/or parameterization. 
In a comparative quest, this study examines flexible architectural designs of three foundation models: DOFA \cite{xiong2024dofa}, SatMAE \cite{cong2023satmae}, and Flex (ClimaX-based) \cite{nguyen2023climax}. The driving motivations for this study hinges on the following questions: (1) What is the optimal tokenization approach to handle inherent varying spectral response properties of EO imagery? (2) How best to integrate and foster preservation of information from different bands (channel) or modalities? To enable a fair comparison across SatMAE, Flex, and DOFA architectures, all models are pretrained under the same conditions using a shared Sentinel-2 dataset, consistent model configurations, and identical self-supervised learning strategies. This uniform setup isolates architectural contributions from other variables. For downstream tasks, we follow a standardized evaluation protocol using the GeoBench \cite{lacoste2023geo} benchmark. Classification tasks use linear probing, while segmentation tasks employ a shared decoder (UPerNet \cite{xiao2018unified}) across all models. Model backbones are frozen, and only decoders are trained, ensuring consistency in adaptation across modalities and architectures. 

The main contributions of the study are in establishing:

\begin{itemize}
    \item \textbf{Standardized benchmarking:} We conduct an apples-to-apples comparison of representative geospatial foundation models by unifying pretraining objectives, datasets and evaluation protocols across classification and segmentation tasks using GeoBench datasets.
    \item \textbf{Architectural insights:} We reveal how tokenization and fusion strategies fundamentally shape model robustness and spectral reasoning, highlighting key trade-offs between spectral flexibility and generalization.
    \item \textbf{Flexibility v. Homogeneity trade-offs:} We conduct additional experiments to compare Flex to a standard ViT show that Flex’s modular design improves adaptability to missing or heterogeneous bands, but may underperform in spectrally homogeneous settings—emphasizing the importance of aligning architecture with data diversity.
\end{itemize}


The paper is structured as follows: Section 2 reviews related works on multimodal and remote sensing foundation models. Section 3 outlines the methods, including a description of the three key architectures—Flex, DOFA, and SatMAE—along with details of the Sentinel-2 pretraining dataset and the GeoBench benchmark used for evaluation. Section 4 presents experimental results, covering implementation details, comprehensive comparisons on GeoBench using Sentinel-2 pretraining, and an additional analysis comparing the flexibility of Flex to a standard ViT under MillionAID pretraining. Finally, Section 5 concludes with key takeaways and insights.



\section{Related Works}
Large vision foundation models have become the backbone for generic feature learning with large volumes of unlabeled imagery data. Presently in Earth observation imagery applications, varying scales of vision transformers \cite{dosovitskiy2020vit} are the primary AI architecture for Large Earth computer vision models, owing to their promising transferrable learning abilities across heterogeneous EO modalities. 

\noindent\textbf{Self-supervised learning:} From a self-supervised learning perspective, the models can be categorized based on three learning strategies: masked image modeling, contrastive learning, and hybrid learning. Masked image modeling (MIM)\cite{Dias2024oreole,nguyen2023climax,prexl2024senpamae,irvin2023usat,cong2023satmae,xiong2024ofanet,xiong2024dofa,ioannis2025fomo} has emerged as the most popular pretraining strategy for backbones targeting visual understanding of EO imagery. MIM proceeds by masking portions of an image, such as pixels or patches and training models to predict the missing portions using the visible context. In contrast, Contrastive learning (CL) \cite{prexl2023multimodal-cl,ayush2021geography,feng2023crossmodal-cl} proceeds by mapping those two views of similar image inputs toward near-proximity points in latent space (forcing similar features) while preserving the distance to all other images in the batch (commonly known as negative examples). More recently, hybrid variants have been emerging \cite{astruc2024omnisat,astruc2025anysat} to combine benefits of masked auto encoding techniques and contrastive learning. For instance, MIM-based models tend to perform best for downstream tasks requiring fine-grained dense predictions (e.g., semantic segmentation), while CL excels at zero-shot scenarios under linear probing configurations. 


\noindent\textbf{Self-supervised multimodal learning:} From a multimodal reasoning perspective, tokenization and feature fusion schemes deserve special attention. SatMAE\cite{cong2023satmae} proposed encoding multi-spectral data as groups of bands using distinct spectral positional encodings. This is paired with a MIM-based pretraining strategy performing independent masking across multispectral and temporal groups, shown to outperform  spatially-consistent masking strategies. USat\cite{irvin2023usat} models spectral bands with varying spatial scales from multiple sensors and conducts separate patch projections per channel while performing spectral group pooling – thus pooling across channels of each modality. Extending from SatMAE, its pretraining is enhanced by random spatial masking of each band (wavelength) independently. Also under a MIM pretraining strategy, OFA-Net\cite{xiong2024ofanet} enables reasoning across Sentinel-1, Sentinel-2, aerial images and low resolution imagery by applying separate patch embeddings for each modality. Embeddings are fed to a shared encoder, then paired with separate decoders for each modality. Advancing toward a sensor and data type agnostic architecture, the dynamic one-for-all (DOFA) \cite{xiong2024dofa} model performs tokenization across all channels but with wavelength-conditioned dynamic for patch embedding; it incorporates a shared transformer based hypernetworks for pretraining; it is pretrained via both MIM with a variable number of spectral bands, as well as a distillation loss using an ImageNet(RGB)-pretrained teacher model. The sensor parameter aware masked autoencoder (SenPA-MAE)\cite{prexl2024senpamae} performs MIM pretraining on representations obtained from per channel tokenization, paired with encodings of sensor characteristics (i.e., spectral response and spatial resolution associated with each band). In this way, it offers the flexibility to pretrain on imagery of different satellites with non-matching spectral or geometrical sensor characteristics. FoMo\cite{ioannis2025fomo} processes all of the most common modalities in EO domain with a single, sensor-agnostic architecture. It conducts tokenization of each spectral band for each sensor separately. \citet{ioannis2025fomo} considered two projection setups: a) a single linear projection for all spectral bands; b) a separate linear projection for each spectral band. Their experiments show the scheme using single learning projection performing best. Paired with positional and spectral embeddings, it promotes sensor independence and boosts flexibility to process various input settings and broadens the scope of applicable downstream tasks. OmniSat\cite{astruc2024omnisat} employs an approach combining CL and MIM, with later fusion between modalities. Patches are first passed through with modality-specific encoders, with the resulting features being used for CL between modalities, as well as being fed to a modality combining module that employs cross-attention to fuse features across modalities. Resulting multimodal features are then passed to decoders used for MIM pretraining. AnySat\cite{astruc2025anysat} follows a joint embedding predictive architecture (JEPA) paradigm, extending OmniSat with scale-adaptive spatial encoders. Its reconstructions in latent space allows the model to learn more consistent and semantically meaningful features compared to methods based on pixel space. 

In addition to such strategies specifically introduced for EO data, we note the architectural design of the ClimaX\cite{nguyen2023climax} foundation model introduced for weather and climate. To enable reasoning across highly heterogeneous datasets of different variables and physical groundings, it extends the ViT architecture with novel encoding and aggregation blocks. It separately tokenizes each variable, and then employs cross-attention to perform early-fusion of representations across variables (or modalities). In addition to enabling data-driven aggregation of variables with different groundings, such early-fusion strategy is chosen also to avoid the increased computational complexity that would arise if feeding the ViT blocks with separate sequences for each variable. 




In summary, several variants have been introduced in terms of tokenization and modality fusion. Targeting reasoning across data from multiple sensing modalities and spatial resolutions, OFA-Net, OmniSat, AnySat perform tokenization per modality. In contrast, USat, DOFA, SenPa-MAE, FoMo, and the ClimaX climate model rely on tokenization per channel (or variable, for the case of ClimaX). 

Whether for fusing representations derived from different modalities or channels, fusion strategies can be categorized as: \textit{early-fusion}, where aggregation across modality-/channel-specific embeddings happens prior to feeding ViT blocks responsible for feature extraction; \textit{intermediate-fusion}, where separate modality-/channel-specific sequences are fed to ViT blocks, with fusion occurring after feature extraction. For the cases of early-fusion, variants include pooling (e.g., USat), dynamic convolution-based patch embedding (DOFA), and cross-attention (e.g., ClimaX, OmniSat, AnySat). 

In this paper, we attempt to isolate the impacts of these different tokenization and fusion strategies by focusing on a simplified scenario of analysis across bands from a single multispectral source (e.g., Sentinel-2). This is an important research question as we pursue the development of multimodal foundation models capable of extracting complementary strengths from the different bands or different modalities, while offering flexibility for usage in downstream scenarios where only a subset of information is available. For such investigation, we select SatMAE, DOFA, and a ClimaX-inspired approach as representative models adopting main variants of tokenization and fusion schemes. SatMAE\cite{cong2023satmae} is representative in leveraging prior knowledge about channels in the form of grouping subsets of spectral bands, while also representing a intermediate-fusion approach where sequences from each channel group are separately processed by ViT blocks. In contrast, we consider our own  ClimaX-inspired variant that we name Flex that combines separate tokenization per channel with a purely data-driven early-fusion enabled by cross-attention. Finally, DOFA's wavelength-based dynamic hypernetwork represents an early-fusion approach balancing between being data-driven while leveraging prior knowledge in the form wavelengths.

\section{Methods}
\subsection{Architectures considered}
\subsubsection{Flex}

\begin{figure}[htbp]
  \centering
  \includegraphics[width=1.0\linewidth]{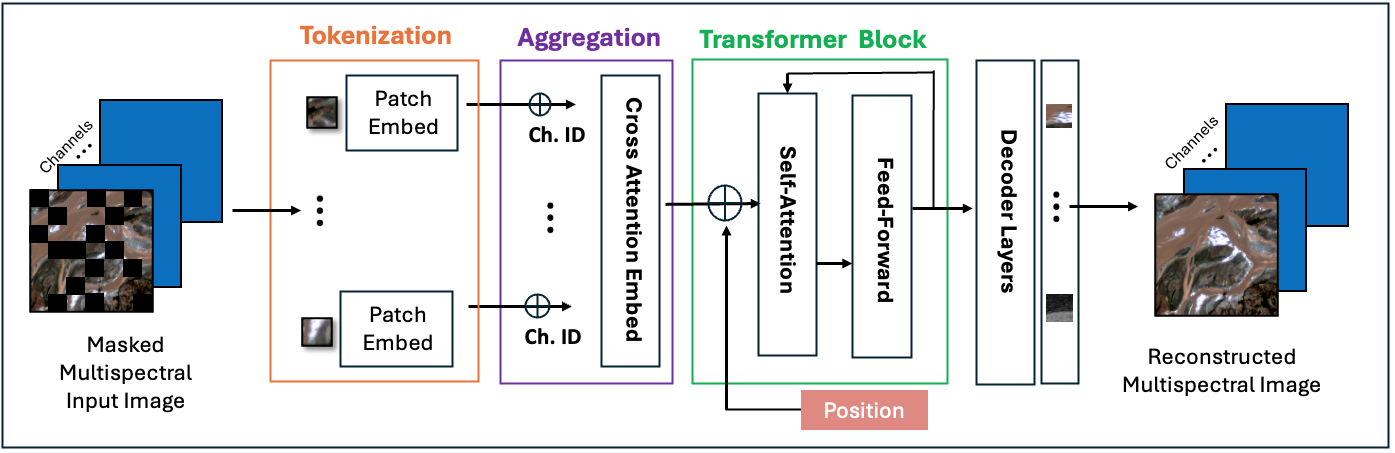}
  \caption{Overview of the Flex architecture. The architecture first tokenizes and embeds each image channel separately using learnable patch embeddings. A cross-attention module, guided by channel embeddings, then fuses information across channels before processing the sequence with Transformer blocks.}
  \label{fig:flex_arch}
\end{figure}

Inspired by the ClimaX architecture \cite{nguyen2023climax}, the Flex architecture differs from a conventional ViT by tokenizing and embedding each channel separately. Let $I \in \mathrm{R}^{C\times H\times W}$ denote an image of $C$ channels and $H\times W$ pixels large. ViTs typically tokenize and encode $I$ into an embedded sequence of shape $(L, D)$, where $L$ is the sequence length (number of patches) and $D$ is the dimensionality of each embedded patch. In contrast, as illustrated in Figure \ref{fig:flex_arch} Flex tokenizes and embeds pixels from each channel separately, with an individual learnable patch embedding for each channel. As such, an embedded sequence of shape $(C\times L, D)$ is first obtained. A process of variable aggregation by means of cross-attention then performs \textit{early-fusion} of representations across channels for each token. To assist this early-fusion step, learnable channel embedding are appended to each token to denote each channel it belongs to (akin to class embeddings). The obtained $(L, D)$-sequence is then fed to subsequent Transformer blocks like conventional ViTs.

\subsubsection{DOFA}

is an architecture trained following the masked image modeling scheme; however, it introduces novel advancement for processing input images with any number of channels and from different modalities. Four key elements underpin the DOFA architecture: i) wavelength-conditioned dynamic patch embedding; ii) shared Transformer network for multimodal pretraining; iii) masked image modeling with a varying number of spectral channel; and iv) a distillation loss for continued multimodal pretraining (which we remove to achieve fair comparison across the three architectures). \textit{Wavelength-conditioned dynamic patch embedding:} this is achieved by a hypernetwork module that generates weights based on central wavelengths of each spectral band. Tokenization is performed across all channels in a patch. Given the patch embedding process is grounded in the wavelength spectrum, this capability enables the hypernetwork to steer the architecture to adapt to novel sensors that get introduced post pretraining. \textit{Shared Transformer network for multimodal pretraining:} DOFA fosters a unified network architecture that learns deep representations across diverse data modalities. To achieve this, it utilizes a shared vision Transformer backbone to universally process heterogeneous data types. \textit{Masked image modeling with a varying number of spectral channel:} The encoding process adapts a masked auto-encoder to reconstruct input data with different number of spectral channels. A wavelength conditioned dynamic decoder layer forms part of the reconstruction workflow. The layer function in conjunction with the wavelength conditioned patch embedding layer. By integrating these dynamic components, the overall DOFA architecture is posed to perform masked reconstruction tasks on imagery inputs with varying spectral bands. 

\subsubsection{SatMAE}

 is a self-supervised learning framework based on masked autoencoders, designed specifically for satellite imagery. By applying a novel masking strategy in joint spatial, temporal, and spectral space—combined with dedicated temporal and spectral positional encodings—SatMAE effectively handles multi-spectral and temporal inputs.
 
SatMAE introduces a group embedding strategy that partitions spectral bands into $G$ groups—typically based on resolution and wavelength similarity—prior to embedding. This design circumvents the limitations of a single patch embedding layer, which assumes uniform inter-channel relationships and may miss fine-grained spectral distinctions. By treating each group $G_j \in \mathbb{R}^{g_j \times H \times W}$ as a separate image and applying distinct patch embeddings per group, the model better captures the unique characteristics of different spectral subsets. $G_j$ is resized to $S_j \in \mathbb{R}^{L \times P^2 g_j}
$ where $L$ is the number of patches and $P$ is the patch dimension. Next, each $S_j$ is embedded into a $D$ dimensional vector $S'_j \in \mathbb{R}^{L \times D}$. Finally, all group embeddings are then concatenated to produce $S' \in \mathbb{R}^{GL \times D}$. This allows the model to represent different groups of channels as token embeddings. 

Moreover, SatMAE also employs independent masking - a scheme whereby each image in a temporal or spectral sequence is patchified separately, and a random subset of patches is masked per image. Hence, the same spatial locations are not necessarily masked across the different images. Instead, a fixed masking ratio (e.g., 75\%) is applied across the full sequence of patch tokens, or per image. Although, such technique can allow a model to infer missing patches if images are highly correlated (eg. video frames), the wider temporal gaps in remotely sensed satellite imagery make it harder to do so.

\subsection{Pretraining dataset}\label{sec:s2_dataset}


\begin{figure}[htbp]
  \centering
  \includegraphics[width=1.0\linewidth]{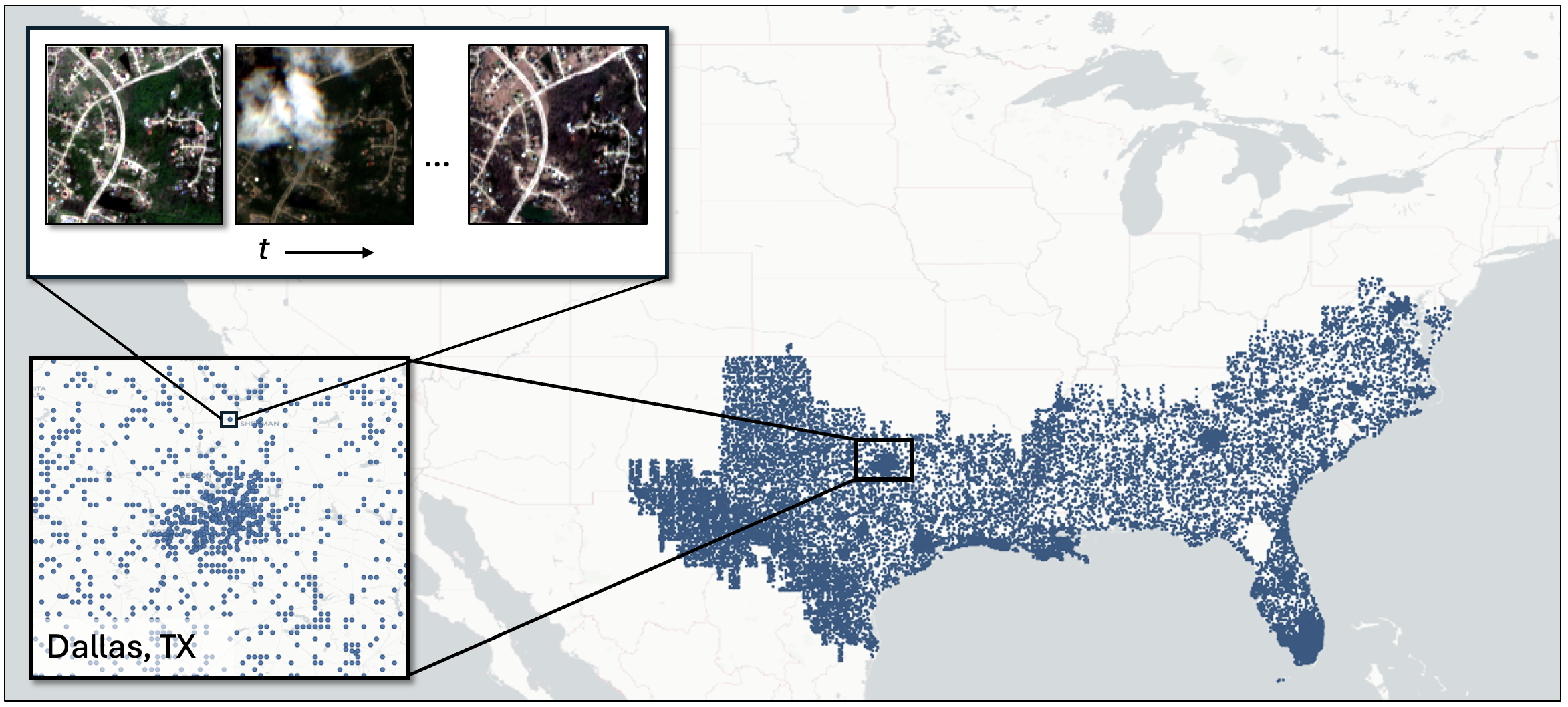}
  \caption{Spatial distribution of sampled points across the southeastern United States used for pretraining. Sampling accounts for environmental and geographic diversity, using criteria such as biomes, climate zones, population density, and land cover. Insets highlight denser sampling in urban areas (e.g., Dallas, TX) and a representative point with its associated Sentinel-2 multi-temporal imagery.}
  \label{fig:my_figure}
\end{figure}

The pretraining dataset for this study, focused on the southeastern United States, with imagery acquired during the year 2021. This region was chosen due to its higher exposure to extreme weather events such as hurricanes and tropical storms, making it well-suited for supporting a wide range of climate-related downstream applications. The year 2021 experienced significant climate activity in the U.S., with 20 confirmed weather and climate disasters, each causing over \$1 billion in damage \cite{smith20222021}. These included several severe storms, floods, and four tropical cyclones, making 2021 a relevant period for capturing diverse environmental conditions. 

Building on \cite{arndt2024towards}, and aiming to create a diverse, representative, and temporally rich dataset, we sampled 22,549 points over the southeast US, based on criteria such as biomes, biogeographic realms, climate zones, population density, and land cover. These points served as anchors for extracting 128×128 pixel tiles from Sentinel-2 imagery, ensuring broad environmental and geographic coverage. To minimize obfuscation due to clouds, we filtered out all scenes with more than 20\% cloud cover. On average, each point was observed 3 to 4 times per month. This consistent temporal sampling across diverse locations captures rich seasonal and phenological variation, contributing to the temporal context. All selected spectral bands were resampled to a common spatial resolution of 10 meters. The bands included in our study were: AOT (Aerosol Optical Thickness), B02 (Blue), B03 (Green), B04 (Red), B08 (Near Infrared), WVP (Water Vapor Pressure), B05 (Vegetation Red Edge 1), B06 (Vegetation Red Edge 2), B07 (Vegetation Red Edge 3), B11 (Shortwave Infrared 1), B12 (Shortwave Infrared 2), B8A (Narrow Near Infrared), SCL (Scene Classification), B01 (Coastal Aerosol), and B09 (Water Vapor). In total, the dataset comprises of 854,200 image tiles, offering both geographic and temporal richness to support the pretraining of the geospatial foundation model.

\subsubsection{Multimodal dataset over CONUS}

\begin{figure}[htbp]
  \centering
  \includegraphics[width=1.0\linewidth]{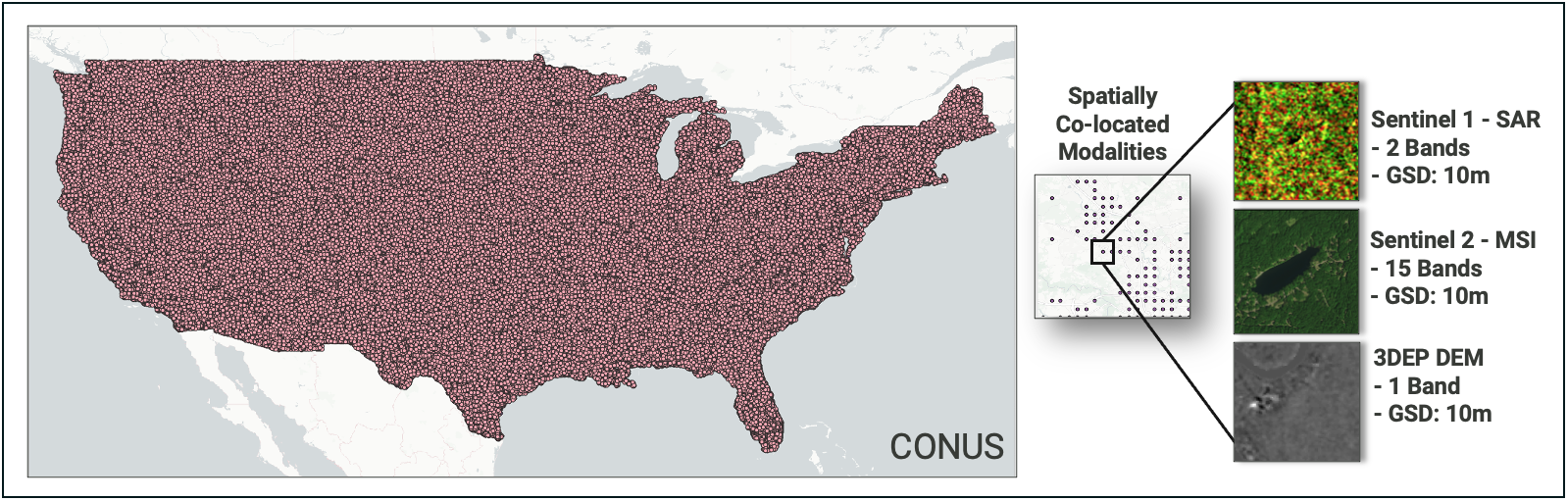}
  \caption{Spatial distribution of sampled points across the Continental United States (CONUS) used for assembling spatially co-located multimodal data. The inset highlights a representative point, showcasing spatial alignment between Sentinel-1 SAR (2 bands), Sentinel-2 MSI (15 bands), and 3DEP DEM (1 band) imagery, all at a 10 m ground sampling distance (GSD).}
  \label{fig:my_figure}
\end{figure}

As part of a broader effort to expand toward multimodal geospatial pretraining, we create a new SatCAMELSH multimodal dataset over CONUS. SatCAMELSH dataset is scheduled for open source release in an upcoming manuscript (the DOI and repository link will be updated in final print of this paper). This dataset serves as an initial step toward a more extensive multimodal corpus and provides a foundation for targeted analyses and ablation studies that complements the pretraining case study.
In constructing the dataset, we extend the sampling strategy to select 250,000 geographically distributed points based on the criteria described previously, ensuring broad representativeness. From this initial pool, we curate $\sim 228$k points that support uniform spatial and temporal coverage. For each selected point, we acquire Sentinel-1 SAR GRD (IW mode) and Sentinel-2 L2A Surface Reflectance imagery for the year 2021. Sentinel-1 scenes undergo geocoding, after which both Sentinel-1 and Sentinel-2 imagery are stacked to produce multiband GeoTIFFs. In parallel, we obtain the 3DEP digital elevation model data at 1/3-arc-second (~10 m) resolution across CONUS, and use the subset of sampled points that spatially overlap the 3DEP coverage to ensure complete multimodal triplets. All three modalities are subsequently tiled into 128 × 128-pixel patches centered on the sampled points. To enforce consistent seasonal sampling, we retain points for which at least one Sentinel-1 and one Sentinel-2 observation are available in each season (Winter, Spring, Summer, Fall), yielding four temporally distributed multimodal samples per point. Sentinel-1 and Sentinel-2 observations are paired such that their acquisition dates differ by no more than 30 days, with each image used only once to ensure unique S1–S2 pairs. Because topography is largely static, a single 3DEP tile is associated with each point. Combining the seasonal S1–S2 pairs with their corresponding 3DEP tiles results in a total of 872,504 multimodal S1–S2–3DEP triplets across CONUS, amounting to a total of 1.96 million image tiles. In this manuscript, we perform experiments using S1-S2 tiles.

\subsection{GeoBench}

\noindent\textbf{GeoBench \cite{lacoste2023geo}} (General Earth Observation benchmark) is a standardized benchmark suite for evaluating geospatial foundation models across a diverse range of Earth observation tasks, including classification and segmentation. It comprises curated, permissively licensed datasets spanning multiple sensor modalities and spatial resolutions, all paired with expert-validated labels. To ensure a more balanced and meaningful benchmark, large datasets were randomly subsampled to better reflect data-scarce downstream tasks and reduce evaluation bias. Additionally, class imbalance was mitigated by uniformly subsampling overrepresented classes, ensuring that model performance reflects genuine pretraining quality rather than the exploitation of class distributions.

In this work, we first focus on a subset of GeoBench datasets that rely exclusively on Sentinel-2 multi-spectral satellite imagery. Specifically, we select the classification benchmarks m-bigearthnet, m-brick-kiln, and m-eurosat, as well as the segmentation tasks m-cashew-plantation and m-SA-crop-type. By restricting our evaluation to tasks using only multispectral Sentinel-2 data, we enable a consistent assessment framework within a single sensor modality, allowing for clearer insights into model performance. Moreover, we further extend our experiment to include the m-so2sat Sentinel-1/Sentinel-2 multi-modal downstream classification dataset. Below, we summarize the original preprocessing steps for each selected dataset.

\textbf{m-bigearthnet:} The Sentinel-2 preprocessing in BigEarthNet-MM involved selecting 125 tiles with less than 1\% cloud cover acquired between June 2017 and May 2018 across 10 European countries. These tiles were atmospherically corrected using ESA’s sen2cor tool to generate Level-2A products, during which the cirrus band (band 10) was dropped. The corrected tiles were then divided into 590,326 non-overlapping image patches at varying resolutions based on band type: 120×120 pixels for 10m bands, 60×60 for 20m bands, and 20×20 for 60m bands. Seasonal representation was considered, resulting in patches distributed across autumn, winter, spring, and summer. A manual quality check identified 70,987 patches fully covered by snow, cloud, or shadow, which were flagged accordingly. \cite{sumbul2021bigearthnet}

\textbf{m-cashew:} To ensure consistent spatial resolution across bands, six 20m Sentinel-2 bands (red-edge and SWIR) were upscaled to 10m using nearest neighbor interpolation, aligning them with the native 10m RGB and NIR bands. For cloud masking, a Landsat-based cloud score algorithm was adapted for Sentinel-2, leveraging five bands (RGB, Aerosol, Cirrus) and two spectral indices (NDMI and NDSI) to detect bright and moist cloud pixels. Since Sentinel-2 L2A lacks a cirrus band, data from L1C was used to compute the cloud mask, which was then applied to L2A imagery. \cite{yin2023mapping}

\textbf{m-brick-kiln:} The authors retrieved all 13 Sentinel-2 bands (B1–B12) from the Earth Engine catalog, using imagery collected between October 2018 and May 2019 to align with the period of ground truth kiln location identification. They compiled a dataset consisting of 6,329 positive and 67,284 negative examples, and divided it into training, validation, and test splits with an 80-10-10 ratio while preserving class proportions. \cite{lee2021scalable}

\begin{table}[ht]
\caption{Summary of GEOBench datasets used in this study.}
\renewcommand{\arraystretch}{1.3}
\rowcolors{2}{gray!10}{white}
\resizebox{\linewidth}{!}{
\begin{tabular}{L{3.2cm} L{2.5cm} L{1.8cm} cccccccc}
\toprule
\textbf{Name} & \textbf{Task} & \textbf{Image Size} & \textbf{Classes} & \textbf{Train} & \textbf{Val} & \textbf{Test} & \textbf{Bands} & \textbf{RGB Res} & \textbf{Sensors} \\
\midrule
m-bigearthnet & Multilabel Classification & 120$\times$120 & 43 & 20000 & 1000 & 1000 & 12 & 10m & Sentinel-2 \\
m-brick-kiln & Classification & 64$\times$64 & 2 & 15063 & 999 & 999 & 13 & 10m & Sentinel-2 \\
m-eurosat & Classification & 64$\times$64 & 10 & 2000 & 1000 & 1000 & 13 & 10m & Sentinel-2 \\
m-cashew-plantation & Semantic Segmentation & 256$\times$256 & 7 & 1350 & 400 & 50 & 13 & 10m & Sentinel-2 \\
m-SA-crop-type & Semantic Segmentation & 256$\times$256 & 10 & 3000 & 1000 & 1000 & 13 & 10m & Sentinel-2 \\
\midrule
m-so2sat & Classification & 32x32 & 17 & 19992 & 986 & 986 & 18 & 10m & S1/S2 \\

\bottomrule
\end{tabular}
}
\end{table}

\textbf{m-eurosat:} To preprocess the EuroSAT dataset, the authors selected 27,000 image patches (64×64 pixels) from Sentinel-2 satellite imagery, covering 10 land use and land cover classes that are visually distinguishable at 10m resolution and commonly found in the European Urban Atlas. All images were manually verified to remove mislabeled samples and those obscured by snow or ice, though samples with color distortions from the lack of atmospheric correction were retained to help models learn real-world variability. The dataset, based on open and frequently updated Sentinel-2 imagery, is also released in a geo-referenced format to support practical Earth observation applications. \cite{helber2019eurosat}

\textbf{m-SA-crop-type:} To prepare the dataset, the authors processed Sentinel-1 and Sentinel-2 time series data alongside high-resolution, daily Planet Fusion imagery (3m, RGB+NIR) to support crop type mapping in Brandenburg, Germany. The Sentinel-2 data was harmonized with Planet imagery using the HLS (Harmonized Landsat Sentinel-2) standard and included as part of a cloud- and shadow-free, analysis-ready dataset. Crop field boundaries were sourced from precise, self-reported EU agricultural subsidy data, with fields smaller than 1km² excluded and raw crop IDs aggregated into nine major crop type classes. Spatially distinct 24km$\times$24km train/test tiles from different years were used to promote generalization across time. \cite{esa_fusion_competition}

\textbf{m-so2sat:} \citet{zhu2020so2sat} created this LCZ42 dataset to meet the need for high-quality global Local Climate Zone (LCZ) training data. The authors manually labeled polygons in 42 major cities and 10 smaller regions across all continents (except Antarctica), then projected these labels onto co-registered Sentinel-1 and Sentinel-2 imagery. This process produced 400,673 paired Sentinel-1/Sentinel-2 patches, each labeled with an LCZ class. Dataset patches, which correspond to 320 × 320 m on the ground, are represented as 32 × 32 pixels with 18 spectral bands, at 10 m resolution. The Geobench version contains around 22K samples.

\section{Experiments}

We conducted two sets of experiments. In the first set, SatMAE, Flex, and DOFA are pretrained on the Sentinel2(S2)-only Southeast United States of America (SEUSA) dataset described in Section \ref{sec:s2_dataset}, and evaluated across the S2-only GeoBench's datsaets: \textit{m-bigearthnet, m-eurosat, m-brick-kiln} for classification, \textit{m-cashew, m-SA-crop-type} for segmentation. In the second set, the three models were instead pretrained on the pairs of Sentinel(S1)+S2 images composing the described CONUS dataset, and evaluated on the S1+S2 \textit{m-so2sat} classification dataset.

\subsection{Implementation details}
\noindent\textbf{Pretraining:} Our analysis targets an ``apples-to-apples'' comparison of architectural designs targeting multimodal reasoning and flexibility to support downstream use cases that drops subsets of multispectral bands. To isolate architecture-dependent performance effects, we pretrain SatMAE, Flex, and DOFA on the SEUSA dataset (described in Sec.~\ref{sec:s2_dataset}), using masked autoencoding (MAE) as the SSL strategy. All models are set to equivalent of ViT-Base configuration, and are pretrained using the same hyperparameters, optimizer, and schedules/duration. For DOFA, this means removing the distillation loss originally paired with MAE reconstruction in its original paper \cite{xiong2024dofa}. For SatMAE, each channel remains independently masked as suggested in \cite{cong2023satmae}, with groups defined as \textit{RGBNIR} $([B02, B03, B04, B08])$, \textit{RedEdge}$([B05, B06, B07, B8A])$, \textit{SWIR} $([B11, B12])$ for S2, and additional group \textit{SAR}$([VV,VH])$ group for S1. For both sets of experiments, models are pretrained for 200 epochs, using image sizes of $128\times128$px, patch-size $16$, mask ratio $0.75$, and normalized pixel values of each masked patch as reconstruction target. AdamW is set as optimizer ($\beta=(0.9, 0.95)$), with weight decay $0.05$, with a cosine annealing learning rate (LR) scheduler ($10$ epochs of warmup, base LR $3e-4$), and batch-size of 256 per GPU using $8$ NVIDIA A100s 80GB (i.e., global batch size of $2048$).

\noindent\textbf{Downstream adaptation:} The backbone architectures include DOFA-base encoder, Flex encoder and SatMAE encoder all with an embedding dimension of 768. We use a patch size of $16\times16$ and a batch size of $16$ for all experiments. We use the AdamW optimizer and scan both optimal learning rate and weight decay values. The downstream experiments are run for a maximum of $50$ epochs. We follow terratorch-iterate \cite{gomes2025terratorch} settings applied to GeoBench datasets where standard data augmentation is applied to training images using random horizontal and vertical flips (each with 50\% probability), followed by resizing to $224\times224$ pixels and conversion to tensors. Validation and test images are only resized to $224\times224$ and converted to tensors, ensuring consistent input dimensions without augmentation. To enable a more robust comparison, we used the Terratorch iterate \footnote{\url{https://github.com/terrastackai/iterate}} framework to perform both hyperparameter search and multi-seed repetitions. Following recommendations in \cite{lacoste2023geo}, each linear probing/UperNet experiment was given a budget of $16$ trials, after which the best hyperparameter configuration was rerun $10$ times with different random seeds for initialization. The hyperparameter search employed Bayesian optimization over learning rates in the range $1e–4$ and $1e-1$, batch sizes of $(16, 32)$, and weight decay values between $0$ and $0.1$. All downstream experiments run for $50$ epochs.  Downstream classification tasks follow a linear probing scheme, using a single linear layer as a decoder. UperNet \cite{xiao2018unified} decoder is used for the semantic segmentation tasks, leveraging multi-scale features extracted blocks $2, 5, 8,$ and $11$ of each corresponding backbone. In contrast to the early-fusion performed by Flex and DOFA, SatMAE relies on intermediate-fusion after features are extracted from each group channel. Since \cite{cong2023satmae} does not detail how this is performed for segmentation experiments, we opt for a mean average pooling across channel-groups at patch-level, thus obtaining feature embeddings of same dimensionality/size as Flex and DOFA to feed corresponding linear probing heads and UperNet. All downstream experiments are run with \textit{frozen} model backbone,  training only the corresponding layers of the decoder. 



\subsection{Sentinel-2 pretraining with downstream evaluation across different band subsets}

\begin{table}[ht]
\caption{Downstream task performance of foundation model backbones (SatMAE, DOFA, and Flex) evaluated on GeoBench datasets under 4 (R, G, B, NIR), 6 (R, G, B, Red Edge 1, Red Edge 2, NIR), 8  (R, G, B, Red Edge 1, Red Edge 2, Red Edge 3, NIR, Narrow NIR), and 10-band (R, G, B, Red Edge 1, Red Edge 2, Red Edge 3, NIR, Narrow NIR, SWIR1, SWIR2) spectral configurations. For each dataset and band setting, the best-performing model is shown in \textbf{bold}, the second-best is \underline{underlined}, and the third is left unformatted.}
\renewcommand{\arraystretch}{1.3}
\small
\rowcolors{2}{gray!10}{white}
\resizebox{\linewidth}{!}{
\begin{tabular}{L{2.5cm} L{2.8cm} L{2.2cm} c c c c}
\toprule
\textbf{Task} & \textbf{Dataset} & \textbf{Backbone} 
& \makecell{\textbf{4 Bands} \\ \textit{B02-B04,B08}}
& \makecell{\textbf{6 Bands} \\ \textit{+B05+B06}}
& \makecell{\textbf{8 Bands} \\ \textit{+B07+B8A}}
& \makecell{\textbf{10 Bands} \\ \textit{+B11+B12}} \\
\midrule
Classification & m-BigEarthNet (m-F1) & \textit{SatMAE} & \textbf{52.1$\pm$0.22} & \underline{50.4$\pm$0.30} & \textbf{52.2$\pm$0.43} & \underline{52.6$\pm$0.36} \\
& & \textit{DOFA} & \underline{49.3$\pm$0.23} & \textbf{50.9$\pm$0.27} & \underline{51.9$\pm$0.32} & \textbf{52.8$\pm$0.41} \\
& & \textit{Flex} & 44.3$\pm$0.26 & 45.3$\pm$0.38 & 45.9$\pm$0.31 & 51.2$\pm$0.44 \\
\cmidrule(lr){2-7}
& m-brick-kiln (OA) & \textit{SatMAE} &\textbf{95.7$\pm$0.27} & \underline{93.6$\pm$0.35} & \underline{94.0$\pm$0.70} & \underline{93.9$\pm$0.43} \\
& & \textit{DOFA} &\underline{95.1$\pm$0.17} & \textbf{95.7$\pm$0.24} & \textbf{96.5$\pm$0.15} & \textbf{96.4$\pm$0.36} \\
& & \textit{Flex} & 89.1$\pm$0.53 & 88.4$\pm$0.59 & 88.9$\pm$0.51 & 92.7$\pm$0.16 \\
\cmidrule(lr){2-7}
& m-Eurosat (OA) & \textit{SatMAE} & \textbf{93.3$\pm$0.50} & \textbf{91.1$\pm$0.39} & \textbf{91.0$\pm$0.49} & \textbf{91.8$\pm$0.48} \\
& & \textit{DOFA} & \underline{89.3$\pm$0.74} & \underline{90.8$\pm$0.19} & \underline{88.4$\pm$1.11} & 88.5$\pm$0.74 \\
& & \textit{Flex} & 85.3$\pm$0.64 & 85.8$\pm$0.74 & 83.2$\pm$0.43 & \underline{89.6$\pm$0.46} \\
\midrule
Segmentation & m-Cashew (mIoU) & \textit{SatMAE} & \textbf{59.0$\pm$3.71} & \textbf{61.1$\pm$2.33} & \textbf{57.8$\pm$2.45} & \underline{60.6$\pm$3.68} \\
& & \textit{DOFA} & 44.5$\pm$9.28 & \underline{51.2$\pm$2.91} & \underline{50.1$\pm$5.99} & 52.0$\pm$4.51 \\
& & \textit{Flex} & \underline{47.4$\pm$6.08} & 48.2$\pm$6.25 & 46.5$\pm$10.41 & \textbf{60.7$\pm$1.37} \\
\cmidrule(lr){2-7}
& m-SA-Crop (mIoU) & \textit{SatMAE} & \textbf{27.2$\pm$0.81} & \textbf{26.1$\pm$0.69} & \textbf{27.9$\pm$1.04} & \underline{29.0$\pm$0.81} \\
& & \textit{DOFA} & \underline{24.1$\pm$0.68} & \underline{25.8$\pm$0.75} & \underline{27.5$\pm$0.61} & \textbf{30.2$\pm$0.82} \\
& & \textit{Flex} & 21.8$\pm$0.67 & 20.5$\pm$0.81 & 23.3$\pm$0.91 & 28.6$\pm$0.68 \\
\bottomrule
\end{tabular}

}
\label{tab:main}
\end{table}

\begin{figure}[h]
    \centering
    \includegraphics[width=1\linewidth]{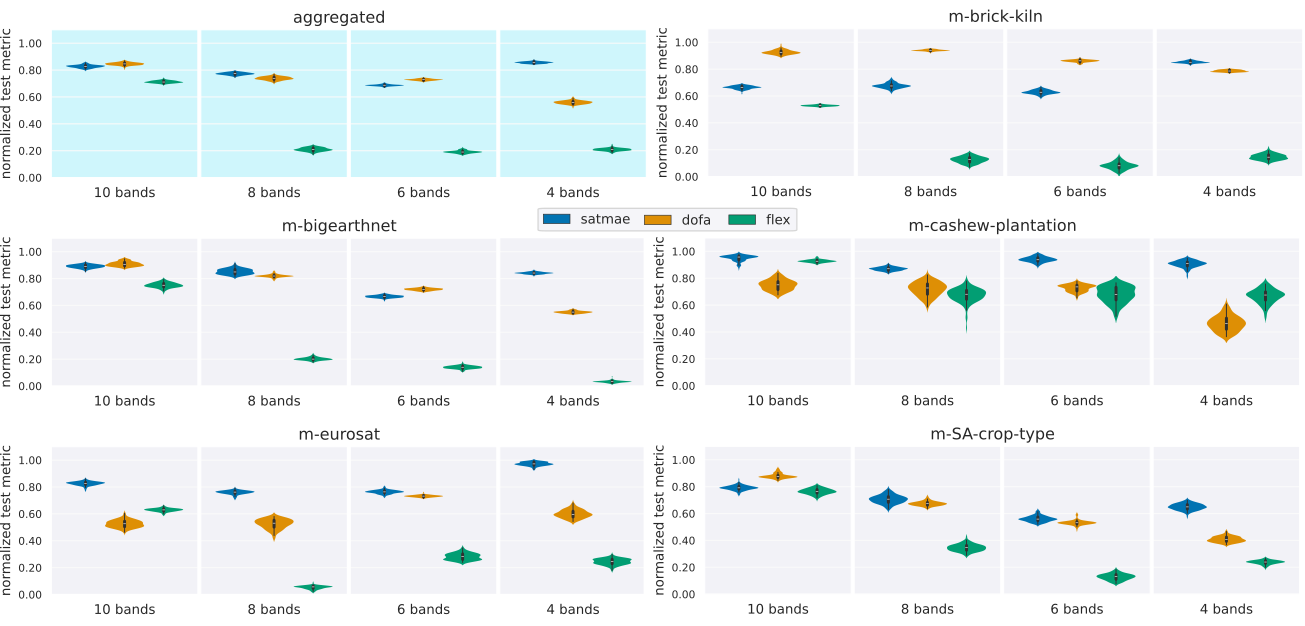}
    \caption{Normalized accuracies of SatMAE, DOFA, Flex across different configurations (higher is better). Following \cite{lacoste2023geo}, violin plots are obtained from bootstrap samples of normalized IQM. }
    \label{fig:drops}
\end{figure}
\begin{figure}[h]
    \centering
        \includegraphics[width=0.9\linewidth]{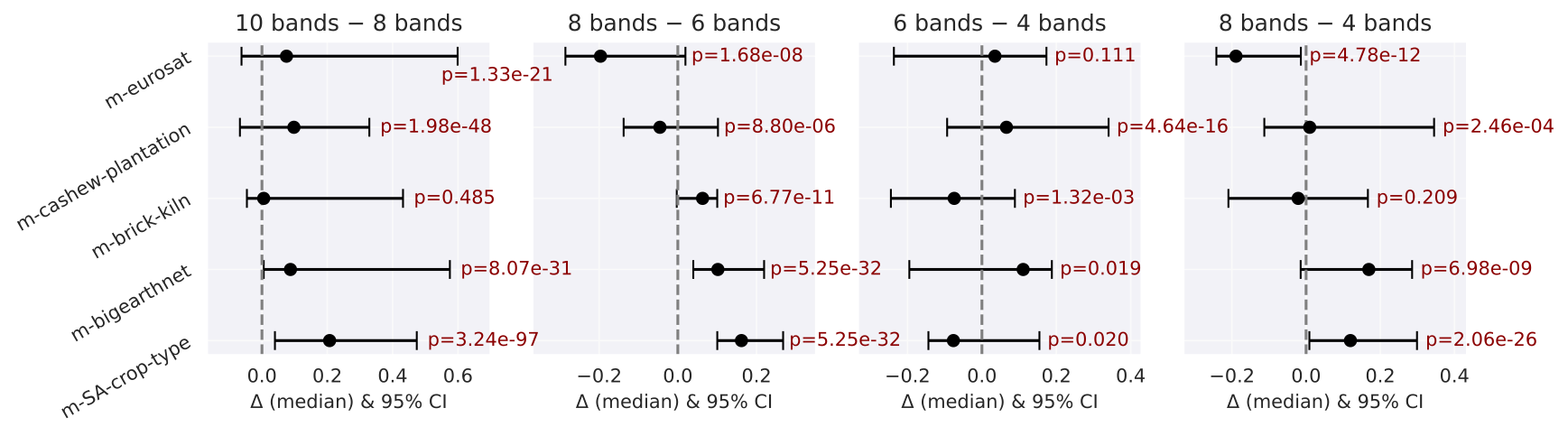}
    \caption{Statistical comparison between band configurations for aggregate performance across models for each dataset. Black points indicate median $\Delta$ values from bootstrap samples, horizontal bars show the $95\%$ bootstrap confidence intervals, and red text indicates the Mann–Whitney U test p-values. The dashed vertical line at zero represents no difference; positive values indicate better performance of configuration A, negative values indicate better performance of configuration B.}
    \label{fig:datadrops}
\end{figure}

\noindent\textbf{Experimental Setup} We conduct four sets of downstream experiments using different datasets and model architectures with varying band configurations. These include:
\begin{enumerate}
    \item \textit{10-bands} experiment, using B02 (BLUE), B03 (GREEN), B04 (RED), B05 (RED\_EDGE\_1), B06 (RED\_EDGE\_2), B07 (RED\_EDGE\_3), B08 (BROAD\_NIR), B8A (NARROW\_NIR), B11 (SWIR\_1), B12 (SWIR\_2) bands;
    \item \textit{8-bands} experiment,  with B02, B03, B04, B05, B06, B07, B08, B8A bands (i.e., SWIR bands B11-B12 are dropped);
    \item 6-bands experiment, with B02, B03, B04, B05, B06, B08 bands (i.e., B07 and B8A dropped);
    \item 4-bands experiment, with B02, B03, B04, B08 (i.e., B05-B06 dropped, with visible spectrum and Near Infrared remaining). 
\end{enumerate}

Table \ref{tab:main} reports the resulting mean and standard deviation for each experiment. Following GeoBench's recommendations \cite{lacoste2023geo}, Figure \ref{fig:drops} summarizes performances in terms of violin plots generated based on bootstrapping over observed interquantile means (IQM), with results normalized to enable aggregating metrics across multiple configurations. In our case, we normalize using lowest and highest metric values achieved across all (model, bands) configurations for each downstream dataset.


\noindent\textbf{Detailed discussion.} Overall, SatMAE emerges as most consistent and better performing architecture, outperforming both DOFA and Flex across 13/20 dataset-bands configurations. 
Most importantly, more detailed analyses reveal informative patterns on advantages/challenges associated with each of the three architectural designs. 
For the analyses that follow, we use $\rightarrow$ to denote comparisons between pair of experiments dropping bands (e.g., $8\rightarrow6$ denotes comparison between results for 8-band vs 6-band).


SatMAE presents the most stable performance across band drops, suggesting its intermediate-fusion approach grounded on prior knowledge for channel-grouping provides a better balance between quality of feature extraction and downstream flexibility. While DOFA provides higher metrics at 10 bands for \textit{m-brick-kiln, m-BigEarthNet} and \textit{m-SA-crop}, it also shows the largest $6\rightarrow4$ vulnerability. This indicates DOFA relies on different regions of the spectrum more evenly, and since bands (B07,B8A) are centered near B08 in Sentinel-2's spectral response, its wavelength-aware patch-embedding provides robustness to the $8\rightarrow6$ configuration change. In turn, SatMAE shows performance increases for \textit{m-BigEarthNet, m-brick-kiln}, \textit{m-eurosat} and \textit{m-SA-crop} in $6\rightarrow4$. For \textit{m-brick-kiln} and \textit{m-eurosat}, its performance with 4bands even surpasses its 10 bands counterparts. We conjecture this is a consequence of how the intermediate-fusion relies on a mean pooling across features extracted from the different group channels: once both SWIR and Red-Edge bands are all dropped, the model effectively relies only on features extracted from the visible spectrum.


Flex's performance drops very significantly across all datasets in $10\rightarrow8$ where SWIR is dropped, which suggests Flex's features focus more on SWIR than its counterparts. This is further confirmed by an additional 8-band experiment where we drop \textit{B05+B06} instead of the SWIR \textit{B11+B012} bands: in that case, Flex still provided mean mIoU $59.2\%$ for \textit{m-cashew} and $27.5\%$ for \textit{m-SA-crop}, a much lower drop with respect to its performance for 10 bands. Results reveal this reliance on SWIR to be disadvantageous for most of the evaluated downstream tasks, with Flex largely outperformed by SatMAE and DOFA across most configurations. However, it is largely suited for the \textit{m-cashew} downstream dataset, where Flex's mIoU is tied with SatMAE as the best for the 10bands configuration. 


To assess potential dataset-specific patterns as bands are dropped, using the \textit{Mann-Whitney U test} we compared the aggregated distributions of bootstrapped IQMs across the three models for each configuration of bands. The Mann-Whitney U test is a nonparametric test between two distributions $(A,B)$ that assesses whether distribution A tends to have larger values than B, without assuming normality/symmetry and suitable when estimates come from independent runs. Under a null-hypothesis of no difference in normalized test metrics between the paired configurations, Figure \ref{fig:datadrops} summarizes median, confidence intervals (CI) and \textit{p}-values obtained for each dataset when comparing the $10\rightarrow8$, $8\rightarrow6$, $6\rightarrow4$ band drops. 

For $10\rightarrow8$, although a large CI range is observed due to Flex's influence, the median statistics highlight how normalized scores significantly drop for \textit{m-SA-crop} and \textit{m-cashew} in particular. This suggests SWIR bands are particularly informative for the two downstream tasks, which makes sense as both datasets require distinguishing classes such as crops, mixed/trees, grasslands, and built-up land, and SWIR is correlated to moisture presence. In turn, these bands have negligible influence for \textit{m-brick-kiln}, which focuses on binary classification of built infrastructure. Analysis of $8\rightarrow6$ capture the increase in m-eurosat accuracy for both Flex and DOFA, which may suggest bands $B07-B8A$ are noisy or irrelevant for this dataset, but such early-fusion models are unable to minimize the influence of these bands. 

\textit{M-BigEarthNet} displays the strongest influence of the RedEdge bands $B05-B06$ that are dropped in $8\rightarrow6$, with DOFA and Flex exhibiting high drops in accuracy. Looking at $8\rightarrow4$ as an assessment of influence of the four RedEdge/mid-range bands, for \textit{M-BigEarthNet} significant performance decrease is confirmed to occur across all models. In turn, for \textit{m-eurosat} a pattern of increase is observed. This further highlights SatMAE can better disentangle feature extraction across different regions of the spectrum, maintaining or even increasing its performance when only visible bands and near-infrared are available. In contrast, DOFA is capable to better extract features from mid-range frequencies, but is unable to maintain performance when such bands are unavailable. 


For both \textit{m-eurosat} and \textit{m-brick-kiln} the visible spectrum is highly informative, with all models in average experiencing little performance decrease as other bands are dropped. This is to be expected for\textit{ m-brick-kiln}, as it focuses on binary classification of built infrastructure. In contrast, it is noteworthy how SWIR/Red-Edge bands are highly informative for \textit{m-cashew} and \textit{m-SA-Crop}, which require differentiation between vegetation types. 


\subsection{Sentinel-1 \& -2 pretraining and downstream evaluation}
\begin{figure*}[!ht] 
    \centering
    \begin{minipage}[t]{0.56\textwidth} 
        \centering
        \captionof{table}{Downstream task performance of SatMAE, DOFA, and Flex trained on S1+S2 and evaluated on M-so2sat under varied configurations.}
        \label{tab:so2sat}
        \footnotesize 
        \rowcolors{2}{gray!10}{white}
        \setlength{\tabcolsep}{2.5pt} 
        \begin{tabular}{l c c c c c}
        \toprule
        \textbf{Backbone} & \textbf{S1+S2} & \multicolumn{3}{c}{\textbf{S2 only}} & \textbf{S1 only} \\
        \cmidrule(lr){3-5}
        & & \textbf{10 bands} & \textbf{8 bands} & \textbf{4 bands} & \\
        \midrule
        \textit{SatMAE} & \textbf{49.7$\pm$1.25} & \textbf{49.6$\pm$0.85} & \textbf{49.1$\pm$1.28} & \textbf{50.4$\pm$1.22} & \underline{19.5$\pm$1.58} \\
        \textit{DOFA}   & \underline{47.9$\pm$1.14} & \underline{46.2$\pm$1.59} & \underline{46.9$\pm$1.68} & \underline{45.3$\pm$1.39 } & \textbf{21.0$\pm$1.66} \\
        \textit{Flex}   & 45.6$\pm$1.10 & \underline{46.2$\pm$0.88} & 43.5$\pm$1.92 & 41.5$\pm$2.17 & 17.2$\pm$1.77 \\
        \bottomrule
        \end{tabular}    
    \end{minipage}%
    \hfill
    \begin{minipage}[t]{0.40\textwidth} 
        \centering
        \vspace{0pt} 
        \includegraphics[width=\linewidth]{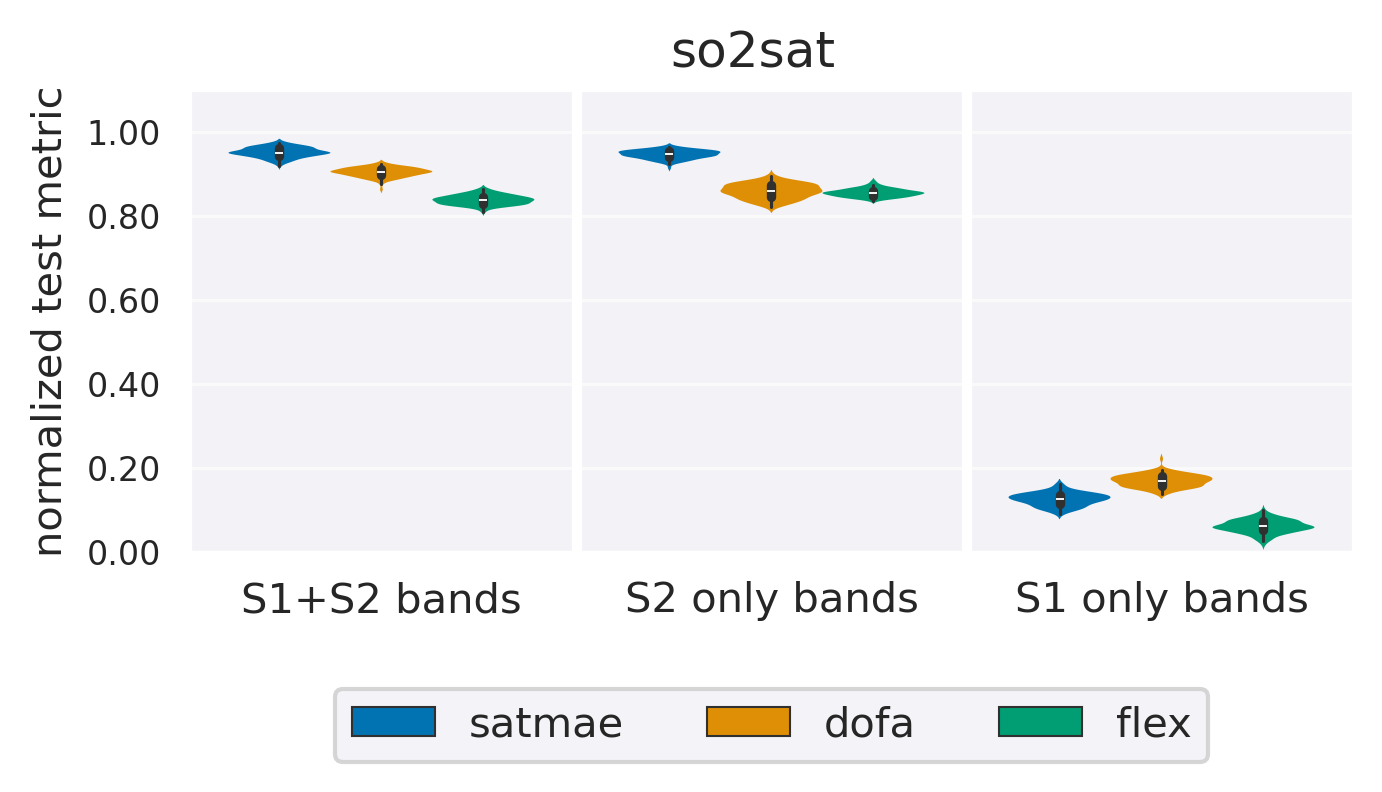}
        \captionof{figure}{Normalized accuracies for each considered model across \textit{S1+S2, S2 only}, and \textit{S1 only} setups.}
        \label{fig:so2sat}
    \end{minipage}
\end{figure*}

In this second set of experiments, we utilized the SatCAMEL multimodal CONUS S1+S2 dataset for pretraining the models and perform downstream adaption to m-so2sat under three configurations: \textit{S1+S2}, \textit{S2 only}, and \textit{S1 only}. Results are summarized in Table \ref{tab:so2sat}, and visualized in {Figure 6}  on boostrapped IQM metrics. Overall, SatMAE performs best for S1+S2 and S2 only configuration, while DOFA performs slightly best when only S1 (SAR) is available. While Flex performs on par with DOFA for S2 only, its accuracy decreases when S2 is combined to S1, indicating an inefficient fusion of modalities when it comes to downstream accuracy. While SatMAE does not perform as well as DOFA in S1 only configuration, it is capable of exploiting its higher quality S2 feature extraction to perform similarly well in S1+S2. 

Moreover, a comparison between \textit{S2 only} results when dropping SWIR bands and RedEdge bands as per the previous Section 4.2 shows that the same patterns observed for the earlier SEUSA/S2-only pretraining dataset do hold in this extended setup of CONUS-wide pretraining on S1+S2 samples. SatMAE continues to exhibit the most stable performance, once again with a slight increase in performance when relying on RGBNIR alone. Meanwhile, Flex's reliance on SWIR bands arise once again, compromising its performance for $10\rightarrow8$ and 4-band only configurations. In turn, DOFA experience a bigger drop when mid-range frequencies are dropped ($10\rightarrow4$).

Regarding potential reasons for its bias towards S2 over S1, we conjecture its MAE pretraining paired with 3 channel groups from S2 and only 1 group for S1 may lead to a higher influence of S2-based losses during optimization, as losses are added across channel groups. From training logs, the losses for each channel group decreased from $[0.291, 0.250, 0.271; 1.075]$ at epoch $1$ vs $[0.104, 0.059, 0.052; 0.635]$ at the final epoch $200$, indicating how: i) S1 data is noisier and hence harder for MAE pretraining; ii) loss minimization was dominated by the first three, S2-based channel groups.

\noindent\textbf{Main takeaways.}
In summary, SatMAE's grouping of channels based on prior knowledge of band relationships shows to be beneficial on both improving robustness to dropping bands, as well as enabling a more balanced feature extraction power across each grouped region of the spectrum. It benefits from the independent masking across bands that it enables, with the caveat observed for S1+S2 experiments that correlated and/or less noisy channels may prevail in pretraining optimization.

Meanwhile, DOFA's wavelength-aware early-fusion enables improved joint feature extraction across the spectrum, providing highest accuracy in several datasets when all 10 Sentinel2 bands are available and improving its accuracy in m-so2sat when considering both S1+S2 as compared to S2-only. However, it presents more severe performance degradation when mid-range (red-edge) bands are not available during downstream adaptation, even in datasets such as m-eurosat where high accuracy can be obtained from RGB+NIR alone (vide SatMAE's performance).

In turn, Flex developed a strong bias for SWIR bands that compromises its accuracy across most datasets. Despite SWIR being of lower importance for many datasets, Flex's early-fusion approach seems to lead to an entangled feature representation that makes it unable to perform well when SWIR is not available. Also performing early-fusion, DOFA seems to strike a slightly better balance by considering wavelength information, in contrast to Flex's cross-attention strategy that is purely data-driven and may succumb to e.g. overly optimizing for noisy channels.

\subsection{Computational Complexity}

Table \ref{tab:model_sizes} provides a summary of model sizes under each configuration. DOFA's larger size is associated with the Transformer/conv blocks composing its dynamic wavelength-based patch embedding. Flex's cross-attention adds approximately 1.7M more parameters than SatMAE, which follows more closely the typical ViT-Base model size. Across the three architectures, DOFA is the most efficient from a compute complexity perspective, achieving the highest throughput (197 images/sec) with the lowest compute (3.36 GMAC) and memory usage (2.07 GB). 
In contrast, SatMAE is the most resource-intensive, as its larger sequence-length leads to 40.26 GMac, 11.36 GB memory (approximately $2\times$ Flex's), and a throughput about $0.28\times$ DOFA's and $0.43\times$ Flex's. Here, GMACs correspond to per-sample complexity (batch size = 1), whereas the memory and throughput comparisons reflect batched inference performance at batch size 32 (10-channel inputs of size 224×224), and should be interpreted accordingly.



\begin{table}[ht]
\caption{Comparison of model parameter sizes and computational efficiency. The table reports parameter counts for classification and segmentation models, as well as computational cost measured in GMACs, inference throughput (images/sec), and peak GPU memory usage (MB). Computation, Throughput and Memory is measured on images of size $224 \times 224$ and 10 spectral bands.}
\renewcommand{\arraystretch}{1.3}
\rowcolors{2}{gray!10}{white}
\centering
\small

\resizebox{\textwidth}{!}{
\begin{tabular}{lccccc}
\toprule
\textbf{Model} & \textbf{Classification model size (params)} & \textbf{Segmentation model size (params)} & \textbf{GMACs} & \textbf{Throughput (img/s)} & \textbf{Peak Memory (MB)} \\
\midrule
DOFA-base      & 111M  & 121M  & 3.36  & 197.19 & 2070.17 \\
Oreole-Flex    & 89.6M & 95.1M & 14.77 & 127.63 & 5193.02 \\
SatMAE         & 87.9M & 93.4M & 40.26 & 55.75  & 11362.34 \\
\bottomrule
\end{tabular}
}

\label{tab:model_sizes}
\end{table}

Overall, the results demonstrate that DOFA offers the best runtime efficiency, Flex provides a middle ground, and SatMAE prioritizes model capacity at the cost of significantly higher compute and memory requirements.

\section{Conclusion}

The insights obtained from presented experiments represents a fertile ground for improving architectural designs toward models that are capable of both extracting informative features from multiple modalities, while also being flexible to enable downstream applications/data where only a subset of spectral bands are available.
 The high-performances of SatMAE across multiple configurations highlight the benefit of: i) leveraging prior knowledge to group bands known to share similar physical groundings, as it yields robustness to cases where only a few bands for each group are absent; ii) separately extracting features from each channel group and then performing intermediate-fusion, rather than extracting features from early-fused representations (as done by DOFA and Flex). 


 From a deployment perspective, our findings highlight how early-fusion architectures, while high-performing, often present performance degradation in cross-sensor (e.g., transitioning a model pretrained on Sentinel-2 to 4-band commercial imagery) or missing-bands scenarios. In contrast, the ``graceful degradation'' observed with SatMAE's grouped-channel architecture provides a critical safety margin for operational pipelines where specific bands may be missing due to atmospheric interference, sensor noise, or regional data gaps. Moreover, it can enable improved performance for tasks that benefit from focusing on specific sub-regions of the spectrum. As our analysis of results per-dataset highlighted, while some tasks/datasets may benefit from, e.g., mid-range Red-Edge bands or higher-frequency SWIR bands, other cases such as \textit{m-eurosat} can be better served by a stronger reliance on the visible spectrum. Overall, later-fusion approaches like SatMAE appear better suited to enable moving toward modular geospatial FMs capable of "graceful degradation" and task-specific optimization, essential for building reliable, cross-constellation pipelines applicable across diverse operational scenarios. 
 


For future advancements, this suggests exploring configurations that pair: i) DOFA-like wavelength-aware embeddings; and ii) SatMAE-style of separate feature extraction and intermediate-fusion for different regions of the spectrum, potentially offering a more efficient versatile and flexible architecture. 




\section{Acknowledgments}

The results and models presented in this work used compute resources from the National AI Research Resource Pilot, with support from NVIDIA, including NVIDIA's DGX Cloud product and the NVIDIA AI Enterprise Software Platform. 

This manuscript has been authored by UT-Battelle, LLC, under contract DE-AC05-00OR22725 with the US Department of Energy (DOE). The US government retains and the publisher, by accepting the article for publication, acknowledges that the US government retains a nonexclusive, paid-up, irrevocable, worldwide license to publish or reproduce the published form of this manuscript, or allow others to do so, for US government purposes. DOE will provide public access to these results of federally sponsored research in accordance with the DOE Public Access Plan (http://energy.gov/downloads/doe-public-access-plan).


\bibliographystyle{ACM-Reference-Format}
\bibliography{sample-base}


\end{document}